\documentclass[twocolumn,letterpaper]{IEEEAerospaceCLS}  % only supports two-column, letterpaper format

\usepackage{float}
\usepackage{amsmath}
\usepackage{graphicx}
\usepackage{upgreek}
\usepackage{tabularx}
\usepackage{multirow}

\usepackage[]{graphicx}    % We use this package in this document
\newcommand{\ignore}[1]{}  % {} empty inside = %% comment

\begin{document}
\title{Integrating Novelty Detection Capabilities with MSL Mastcam Operations to Enhance Data Analysis}

\author{%
Paul Horton\\
Arizona State University\\
Tempe, AZ 85287\\
pahorton@asu.edu
\and 
Hannah R. Kerner \\
University of Maryland\\
College Park, MD 20742 \\
hkerner@umd.edu
\and
Samantha Jacob\\
Arizona State University\\
Tempe, AZ 85287\\
Samantha.Jacob@asu.edu
\and
Ernest Cisneros \\
Arizona State University\\
Tempe, AZ 85287\\
ecisneros@asu.edu
\and
Kiri L. Wagstaff \\
NASA Jet Propulsion Laboratory \\
Pasadena, CA 91109 \\
kiri.l.wagstaff@jpl.nasa.gov
\and
James Bell \\
Arizona State University\\
Tempe, AZ 85287\\
jim.bell@asu.edu
%%%% IMPORTANT: Use the correct copyright information--IEEE, Crown, or U.S. government. %%%%%
\thanks{\footnotesize 978-1-7281-7436-5/21/$\$31.00$ \copyright2021 IEEE}              % This creates the copyright info that is the correct 2021 data.
%\thanks{{U.S. Government work not protected by U.S. copyright}}         % Use this copyright notice only if you are employed by the U.S. Government.
%\thanks{{978-1-7281-7436-5/21/$\$31.00$ \copyright2021 Crown}}          % Use this copyright notice only if you are employed by a crown government (e.g., Canada, UK, Australia).
%\thanks{{978-1-7281-7436-5/21/$\$31.00$ \copyright2021 European Union}}    % Use this copyright notice is you are employed by the European Union.
}

\maketitle

\thispagestyle{plain}
\pagestyle{plain}

\begin{abstract}
While innovations in scientific instrumentation have pushed the boundaries of Mars rover mission capabilities, the increase in data complexity has pressured Mars Science Laboratory (MSL) and future Mars rover operations staff to quickly analyze complex data sets to meet progressively shorter tactical and strategic planning timelines. 
MSLWEB is an internal data tracking tool used by operations staff to perform first pass analysis on MSL image sequences, a series of products taken by the Mast camera, Mastcam. Mastcam consists of a pair of 400-1000 nm wavelength cameras on MSL's Remote Sensing Mast that, among other functions, uses a filter wheel to produce multispectral images by creating a sequence of products at different wavelengths. 
Mastcam's multiband multispectral image sequences require more complex analysis compared to standard 3-band RGB images. 
Typically, these are analyzed by the inspection of false color images created to aid visualization, such as band ratios between different spectral indices that can highlight specific potential mineralogic differences among iron-bearing phases, and decorrelation stretches to enhance the color differences between multiple filters. 

Given the short time frame of tactical planning in which downlinked images might need to be analyzed (within 5-10 hours before the next uplink), there exists a need to triage analysis time to focus on the most important sequences and parts of a sequence. 
We address this need by creating products for MSLWEB that use novelty detection to help operations staff identify unusual data that might be diagnostic of new or atypical compositions or mineralogies detected within an imaging scene. 
This was achieved in two ways: 1) by creating products for each sequence to identify novel regions in the image, and 2) by assigning multispectral sequences a sortable novelty score. 
These new products provide colorized heat maps of inferred novelty that operations staff can use to rapidly review downlinked data and focus their efforts on analyzing potentially new kinds of diagnostic multispectral signatures. 
This approach has the potential to guide scientists to new discoveries by quickly drawing their attention to often subtle variations not detectable with simple color composites.
The products developed in this work have shown promising benefits for integration into mission operations by potentially decreasing tactical operations planning time through guided triage.
\end{abstract}

\tableofcontents

%%%%%%%%%%%%%%%%%%%%%%%%%%%%%%%%%%%%%%
\section{Introduction}
%%%%%%%%%%%%%%%%%%%%%%%%%%%%%%%%%%%%%%
While discovering the unexpected is one of the most exciting parts of research, the process of making a discovery often involves countless hours of sifting through otherwise mundane data. 
Advances in novelty detection systems can help to alleviate this arduous task by enabling researchers to focus their attention on the most interesting parts of their data. 
Novelty is defined as an unexpected occurrence in a sequence based on a data set of typical sequences.
Given what is known to be “normal,” the novelty detection algorithm highlights the most unusual features in an image. 
Novelty detection techniques work by analyzing the commonalities in a data set in order to generalize the structure of the data and pick out anomalies \cite{japkowicz1995novelty}.
When a new observation is presented to a novelty detection system, the system identifies features that differ from the commonalities learned during training \cite{markou2003novelty}.
This is different from traditional classification systems as what constitutes as \textit{novel} is not predefined. 
Instead of classifying samples as \textit{typical} and \textit{novel}, novelty detection systems build a model from \textit{typical} examples that can be used to discover anomalies. 
This makes novelty detection ideal for scenarios where \textit{novel} examples are infrequent or difficult to obtain and \textit{typical} examples are abundant \cite{japkowicz1995novelty}.
Identifying anomalous examples is useful in many application domains such as structural fault detection for aerospace systems by analyzing ambient vibrations \cite{worden1997structural} or identifying brain tumors using MRI images \cite{wang2020brain}.
\begin{figure*}[t!]
\centering
\includegraphics[width=\linewidth]{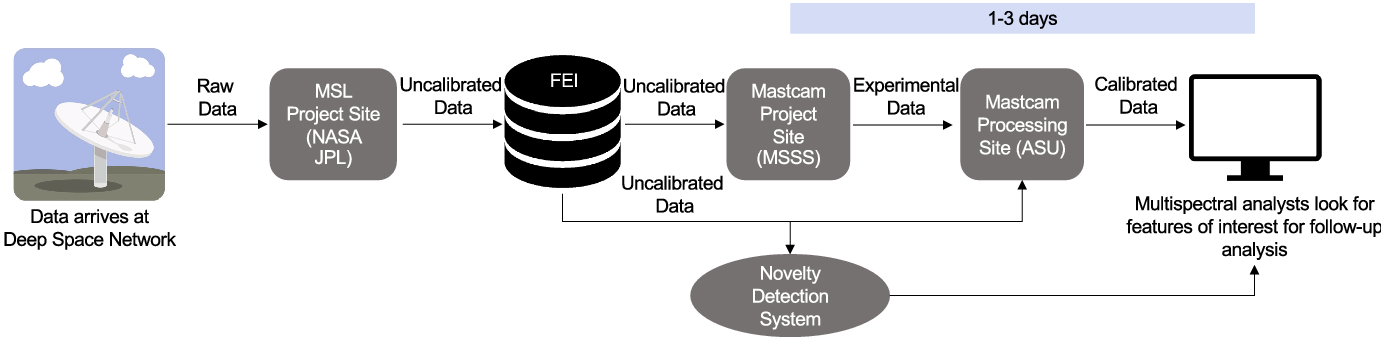}
\caption{\bf{Illustration of the Mastcam image processing pipeline \protect\cite{kerner2020comparison}}}
\label{timeline}
\end{figure*}

Planetary science is another domain where novelty detection proves useful. 
Due to the rapid turn-around required for tactical planning for landed Mars missions, efficient data analysis methods need to be employed to analyze data from scientific instruments \cite{bell2019tactical}.
With high volumes of downlinked data, tactical operations planning teams have to quickly perform ground analysis to meet the ten hour turn-around time for MSL operations to uplink commands \cite{samuels2013preparation}. 
The timeline is even shorter for the upcoming Mars 2020 mission which has a desired five hour turn-around time \cite{wilson2017nasa}.
During tactical planning, the MSL Curiosity rover sends compressed observations through one of three Mars orbiters to the Deep Space Network (DSN) on Earth so that operations team members can use these observations to make plans for the next sol \cite{kerner_multispec}.
Rapid analysis is required for tactical planning as discoveries made outside of the time frame are subject to increased mission resource cost as the rover may need to reverse course to re-visit the target \cite{kerner2020comparison}.
This need for rapid analysis makes systems that can quickly extract information and identify regions of interest in scientific data vital to efficient tactical planning. 

This work primarily focuses on the Mastcam imaging system on the MSL Curiosity rover.
Mastcam consists of a pair of multispectral Charge Coupled Device (CCD) imagers on MSL that are each capable of using an eight-position filter wheel to take red, green, and blue (RGB) images as well as multispectral images in nine bands from 400 to 1100 nm (visible to near-infrared) \cite{bell_mastcam}.
A set of image products from Mastcam is called a sequence.
Seven of the eight landed missions on Mars have employed camera systems capable of multispectral imaging \cite{bell2019tactical}.
Future missions, such as Mars 2020 and Psyche, will also be carrying similar multispectral cameras \cite{bell2016mastcam} \cite{bell_psyche}.
Given that multispectral cameras are so prevalent in planetary exploration, the ability to rapidly detect novel features in multispectral images is beneficial across many missions.

The goal of this work is to operationalize previous work developed for novelty detection for Mastcam.
Kerner et al.~quantitatively compared different novelty detection methods for analysis of multispectral images on Mars \cite{kerner2020comparison}.
With a typical data set, machine learning models including Reed-Xiaoli (RX) detectors, Principal Component Analysis (PCA), Generative Adversarial Networks (GANs), and Convolutional Autoencoders (CAE) were used to produce a measure of how atypical each image or multispectral pixel is in a new sequence. 
Using pixelwise analysis allows these methods to highlight difficult to identify novel regions within an image that may otherwise go undetected by operations staff.
These methods\footnote{https://github.com/JPLMLIA/mastcam-noveltydet} and the data set\footnote{https://zenodo.org/record/3732485} used to evaluate them were made publicly available at publication. 
Multiple models were evaluated to show that certain models perform better for certain novelties than others -- e.g., PCA was better suited for detecting spectral (compositional) novelties than for morphological (shape) novelties. 
Their work provides an in-depth qualitative evaluation between these methods which was used to guide decisions about which methods to use. 
While their work demonstrated the capabilities of the algorithms, it did not evaluate the methods based on their effectiveness in the tactical planning pipeline.
In order to be most useful to operations, these methods need to be automatically run on new data and integrated into tactical analysis workflows to accelerate tactical planning \cite{donahoe2020new}.
New advances in this work include further development of the algorithms and the creation of an implementation strategy for operations. 
Additionally, we analyzed the outputs from the system to determine the usefulness of these methods in comparison to existing tactical planning procedures. 
%%%%%%%%%%%%%%%%%%%%%%%%%%%%%%%%%%%%%%
\section{Related Work}
%%%%%%%%%%%%%%%%%%%%%%%%%%%%%%%%%%%%%%

Novelty detection is a form of anomaly detection that focuses on detecting novel examples in a data set \cite{domingues2019comparative}. 
Given what is known to be typical in a data set, these algorithms find novel examples that may be unknown to the user.
Unlike a classifier, novelty detection systems detect whether an input is similar to examples in the typical training set or if the input is novel \cite{markou2003novelty}.
Novelty detection can be seen as a one-class classification task where a model is trained to describe a data set of typical examples \cite{pimentel2014review}. 
For novelty detection, the training data represent a set of examples that an end user would identify as typical examples. 
When the model is used to infer the novelty of new data, the system calculates how different the new examples are compared to the training set. 
As abnormal examples are not well represented in the training set, novelty detection systems are not able to model abnormalities and thus highlight them as novel. 
This allows novelty detection systems to highlight abnormal data by evaluating how well (or how poorly) it can model the examples. 
These methods can be applied to various domains such as fraud detection based on card activity \cite{oosterlinck2020one}, human verification for websites from mouse and keyboard usage \cite{kim2018keystroke}, fault detection for aerospace systems by analyzing ambient vibrations \cite{worden1997structural} and brain tumor identification using MRI images \cite{wang2020brain}.

This work is based on previous work that developed novelty detection for multispectral images on Mars \cite{kerner2020comparison}.
Figure \ref{timeline} shows the current pipeline for multispectral image processing for operational analysis and where we propose to augment the pipeline with novelty detection systems. 
Calibrated data are often not available during the tight tactical analysis schedule, so uncalibrated data will be used for novelty analysis. 
Current methods for quick analysis involve generating quick look products, such as decorrelation stretches and filter ratios, which help to identify spectral changes in the sequence \cite{gillespie1986color}.
For example, comparing the relative reflectance spectra of two different drill tailings can inform analysts of the similarities and differences in mineral composition \cite{wellington2017visible}.
These quick look products are generated using calibrated data making them unavailable during the tactical analysis time frame.  

Kerner et al.~demonstrated four types of novelty detection systems for multispectral novelty detection: CAEs, GANs, PCA, and RX detectors \cite{kerner2020comparison}. 
All of these methods, except RX, are reconstruction based methods that attempt to ``compress'' and ``reconstruct'' an input to recreate the original image. 
When trained on typical examples, these methods are able to reconstruct normal examples well but are unable to represent novel examples.
The novelty detection system is able to identify novel regions based on the reconstruction error, which is the difference between the input and recreated image.
For Mastcam multispectral sequences, these images have six bands instead of the traditional one (grayscale) or three (RGB). 
It is important to note that Kerner et al.~evaluated these methods to identify novel geology in multispectral sequences, not create a visualization for tactical analysis that maximizes their ability spot hard-to-find features. 
Our goal is to create novelty detection products that could be integrated into actual tactical operations pipelines and evaluate their effectiveness for triaging analyst time in tactical operations.

%While CAEs, GANs, and PCA are good at identifying novelties such as bedrock, veins, and broken rocks, these novelties are typically easily identifiable by looking at the RGB image in the sequence. 

%RX is a distribution-based method that that calculates the distance between a test sample and a calculated background distribution. 
%When using RX to evaluate individual pixels in a sequence, the method has no spatial awareness and performs poorly at detecting novel features that require structural information \cite{kerner2020comparison}.
%However, when RX is used in this manner it performs well on detecting novelties from spectral information, such as drill holes, dust removal tool (DTR) spots, and meteorites. 
%Of the methods tested in Kerner et al.~, RX shows the most viably in a tactical setting as it is typically easier for operations staff to identify structural novelties than spectral ones. 
%%%%%%%%%%%%%%%%%%%%%%%%%%%%%%%%%%%%%%
\section{Data Set}
%%%%%%%%%%%%%%%%%%%%%%%%%%%%%%%%%%%%%%
\begin{figure*}[hbt!]
\centering
\includegraphics[width=\linewidth]{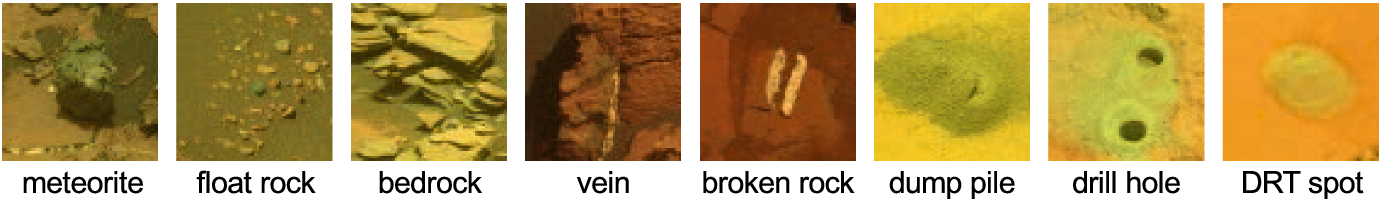}
\caption{\bf{Examples from the eight categories of novel geology in the Mastcam multispectral image data set \protect\cite{kerner_data}} }
\label{NovelCategories}
\end{figure*}

In this work, we created a data set of all multispectral Mastcam sequences based on the pre-processing methods outlined in the labeled Mars multispectral novelty detection data set \cite{kerner_data}.
This previous data set provided images with both \textit{typical} and \textit{novel} labels cropped from the M-100 right eye of Mastcam on MSL between sols 1 to 1666.
Each example in the data set is a 64 by 64 tile with 6 bands corresponding to 6 different filters. 
The tiles were sub-sampled from larger thumbnail images of around 140 by 100 pixels in size. 
Thumbnails were used rather than full resolution sequences as they are among the first available products during tactical planning. 
Additionally, uncalibrated images were used as these data are the most readily available in the tactical analysis timeline. 
To generate the data set, these thumbnails were loaded as single band images in OpenCV before combining them to produce a single, six band, tile \cite{opencv_library}.

While we did not explicitly use the labeled Mars multispectral novelty detection data set in this work, we utilized the novelty classes to verify the system.
The data set provided a set of novel tiles divided into 8 sub-classes: meteorite, float rock, bedrock, vein, broken rock, dump pile, drill hole, and DRT spot, as shown in Figure \ref{NovelCategories}. 
Multispectral analysts from the Mastcam operations team identified novel geology in images using bounding boxes based on operations experience and past publications of high interest targets (e.g., \cite{wellington2017visible}).
Sub-sampled tiles that intersected with these bounding boxes were included in the novel data set.
%Multispectral analysts were tasked with drawing bounding boxes around novel features in thumbnail sequences based on their knowledge of high interest targets supported by literature.
%After cropping the thumbnails into tiles, a set of novel tiles was created from tiles containing these bounding boxes.

% The typical tile set was created by selecting images and image regions not containing a bounding box. 
% The tiles were split at the source thumbnail level into training, validation, and test sets using an 80\%/10\%/10\% split (resulting in 9302 training, 1386 validation, and 856 test tiles) while the novel tiles were used exclusively in the test set. 
% Tiles were split based on their source thumbnail to ensure tiles from the same base sequence were not included in multiple data sets, which would increase the correlation between these datasets which should be as independent as possible.

The goal of this work is to triage tactical analysis time by prioritizing interesting images and highlighting features that are otherwise difficult to identify in multispectral sequences.
While our system should be able to identify novelties such as large veins and broken rocks, which are relatively easy for analysts to identify in RGB or single-band images, highlighting these examples will not reduce tactical analysis time as much as other features that are difficult or impossible to see without detailed analysis of multispectral sequences. 
%they can be easily seen without a novelty detection system.
%Many of the examples in the original training set contain novelties that a tactical operations staff member could easily spot. 
%These examples are visible in the RGB images and do not require a novelty detection system to draw attention to them.
The Kerner et al. \cite{kerner2020comparison} data set did not account for these operational priorities when evaluating novelty detection performance. 
In order to fully evaluate the benefit of our novelty detection system in tactical operations, we created a new data set using the same pre-processing as the Kerner et al. data set. 
Instead of separating the images into typical and novel data sets, we included all multispectral sequences from the left and right eyes of Mastcam in an unsupervised manner.
The images in this data set were pre-processed in the same way as the original data set but kept as full size thumbnails instead of tiles in order to provide detections at the same resolution as they are viewed by operations staff. 
Starting with every thumbnail taken by Mastcam, the data set was filtered by sequences containing 7 different filters: RGB (no filter) and 6 narrow-band spectral filters.
Additionally, sequences containing the calibration tool were omitted. 
This filtering resulted in about 900 total sequences split between the left and right eye. 
The six narrow-band sequences were used as inputs into the algorithm and the RGB is was used for reference.
%%%%%%%%%%%%%%%%%%%%%%%%%%%%%%%%%%%%%%
\section{Methods}
%%%%%%%%%%%%%%%%%%%%%%%%%%%%%%%%%%%%%%
In this work, we employed a the pixelwise RX method for novelty scoring.
Pixelwise RX is a distribution-based method that calculates the distance between each pixel in a test image and a background distribution, which can be visualized at the image level as a heatmap of novelty scores. 
In comparison with other methods used in Kerner et al. \cite{kerner2020comparison}, pixelwise RX is out-performed when identifying images with certain types of novelties such as float rocks, veins, bedrock, and broken rock, but performs better for other novel categories. 
While this may seem like RX is a poor choice for novelty detection, the novelty scores in Kerner et al. were aggregated to image-level scores and did not compare algorithms on the basis of their ability to highlight novel pixels in an image in a way that is useful for tactical planners. 
Pixelwise RX may not have the highest performance for all types of novelty, but its scores and their associated visualizations are relatively simple and interpretable compared to other methods. 
This is an important characteristic in method selection for analysts using novelty detection methods in a tactical operations setting. Thus, we chose to use pixelwise RX for developing novelty-based tactical planning products. 
%Of the novelty detection methods discussed in previous work, algorithms that are sensitive to spectral novelties are most desirable. 
%Spectral novelties are novelties occurring due to unexpected differences in the spectrum of an object and not from the actual shape of the object.
%Methods to detect structural novelties typically pick out features visible in standard RGB images. 
%As the goal of this system is to identify novelties that are not easily observable in the uncalibrated sequences, finding structural novelties is not necessary. 
%While hard to spot structural novelties do exist, we have chosen to focus on spectral novelties as these tend to be more difficult to spot visually. 

In this study, we chose to focus on spectral novelties that are difficult for analysts to spot because they require analysis across multiple filters to find correlations difficult to identify when looking at each image separately. 
%The method chosen to detect spectral novelties is pixel-wise RX as previous work has shown it is best suited for identifying novelties with spectral differences when compared with CAEs, GANs, and PCA \cite{kerner2020comparison}.
While methods not based on novelty detection exist that help identify spectral novelties, they require calibrated data which are often unavailable during the tactical time frame. 

Unlike the other algorithms considered, RX is not a reconstruction based method for detecting novelties.
Instead, RX computes a background distribution from typical examples and compares this distribution to infer the novelty of pixels in new images \cite{reed1990adaptive}.
For novelty detection in Mastcam sequences, the background is computed using the spectrum from each pixel in all sequences in order to generalize the entirety of the data set. 
For a multispectral image with $n$ filters, the background distribution is defined by the $n \times 1$ mean spectrum vector, $\boldsymbol{\mu}$, and an $n \times n$ covariance matrix, $\boldsymbol{\Sigma}$, of all pixels in the training set \cite{guo2016anomaly}.
This provides a mean for each multispectral band and a covariance matrix for the band pairings. 
To infer the novelty in a new image, $\boldsymbol{X}$, an RX score is calculated for each $n \times 1$ pixel vector, $\boldsymbol{x}_i \in \boldsymbol{X}$, using the Mahalanobis distance between the background and the filter response values as shown in Equation \ref{RX_eq}.
\begin{equation}
    \label{RX_eq}
    \text{RX}(\boldsymbol{x}_i) = (\boldsymbol{x}_i - \boldsymbol{\mu})^T \boldsymbol{\Sigma}^{-1} (\boldsymbol{x}_i - \boldsymbol{\mu})
\end{equation}
This score will be referred to as the pixel novelty score as it is a measure of how novel a pixel is relative to the background distribution. 
As this method is pixel based, the dimension size of the sequence does not matter so inferences can be run on sequences of any resolution.
This is particularly useful for Mastcam as the height of each sequence of images varies. 

After computing the RX background distribution using typical sequences, Equation \ref{RX_eq} can be applied to new sequences to identify novel features.
Each sequence can be analyzed individually or relative to other sequences based on their pixel novelty scores. 
To reduce the effect of brightness in the novelty scoring, separate models were created using a data set where the RGB image of the sequence is loaded as a gray scale image and used to divide the other six images in the input. 
Individual sequence analysis is accomplished by visualizing the pixel novelty scores as a heat map. 
Analysts can quickly review these heat maps to identify the most novel regions within a sequence relative to all prior sequences. 
To compare the novelty of different sequences, statistics can be calculated for each sequence and sorted.
This can help prioritize which multispectral sequence to review first. 
%%%%%%%%%%%%%%%%%%%%%%%%%%%%%%%%%%%%%%
\section{Results}
\begin{figure}
\centering
\includegraphics[width=3.25in]{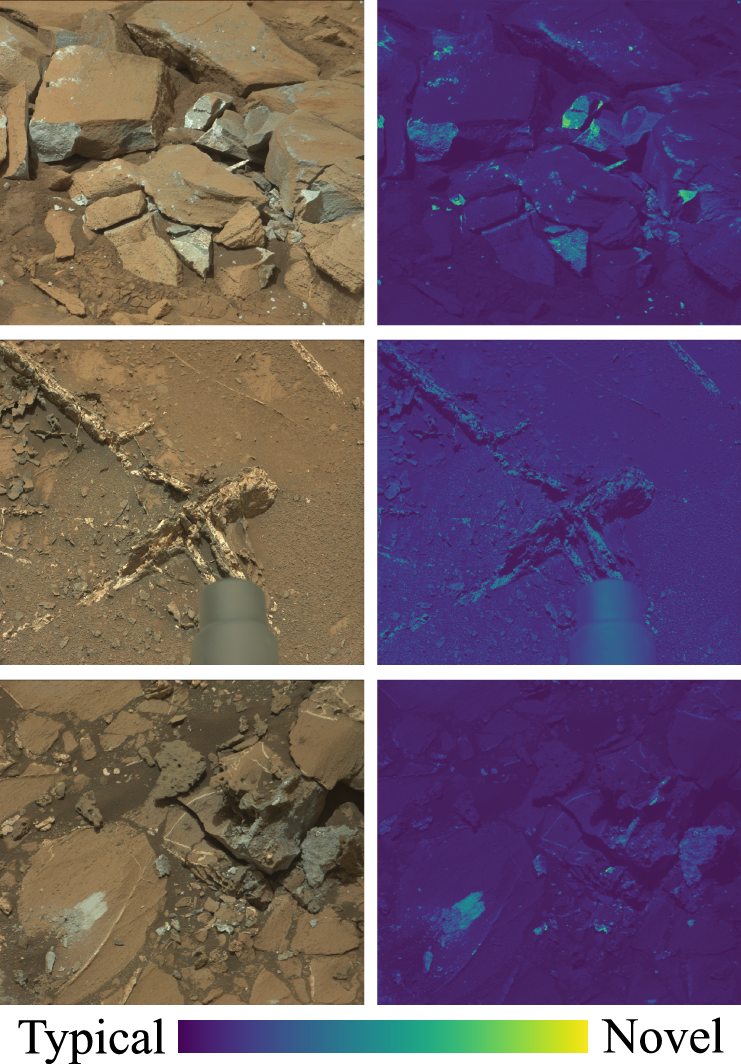}\\
\caption{\textbf{Examples of pixel novelty scores using RX for easy to spot novelties. The left column shows the RGB image from the sequence and the right column shows the output from RX with a color bar ranging from typical to novel. From top to bottom: \mbox{Broken Rock (mcam05168)}, \mbox{Veins (mcam04817)}, \mbox{Dump Pile (mcam04892)}}}
\label{easyNovel}
\end{figure}

%%%%%%%%%%%%%%%%%%%%%%%%%%%%%%%%%%%%%%
%Using the training set created from all the sequences, a background distribution was created for the typical Mastcam multispectral pixel. 
%This background distribution was used to generate pixel novelty scores for each sequence in this same data set.
To assess the system's performance for the novel features in the Kerner et al. data set \cite{kerner_data}, we located the full-size thumbnails of a selection of the novel tiles from each category and verified that the novelties highlighted in the new images were self consistent with previous results in the tiles. 
The novelty detection system using pixelwise RX performs well for highlighting easy-to-spot novelties in the data set. 
Figure \ref{easyNovel} shows examples of some of the easy to catch novelties that the system is able to detect. 
We claim that these examples are relatively easy for analysts to identify because the novelties they highlight can be found through careful analysis of a single RGB image.
In the first example, a broken rock is shown near the center of the image.
The novelty heat map highlights the broken rock well as well other broken rocks in the area as shown by the yellow regions.
The second example shows veins which are highlighted well in its corresponding heat map. 
By just looking at the RGB thumbnail, an untrained analyst could likely spot something novel in the image. 
Finally, the last image shows a dump pile.
This example is clearly visible in the RGB, but would also be easy to identify because the dump pile was created by the rover, so operations staff will undoubtedly be aware of its presence in the sequence. 
While these examples are novel and interesting, they are not the type of novelty that a tactical operations staff member is likely to miss, though highlighting these examples may help to pick them out of a large set of downlinked images. 
%Highlighting these examples may help to pick them out of a large set of downlinked images but the analysis is unnecessary for individual analysis.

\begin{figure*}
\centering
\includegraphics[width=\linewidth]{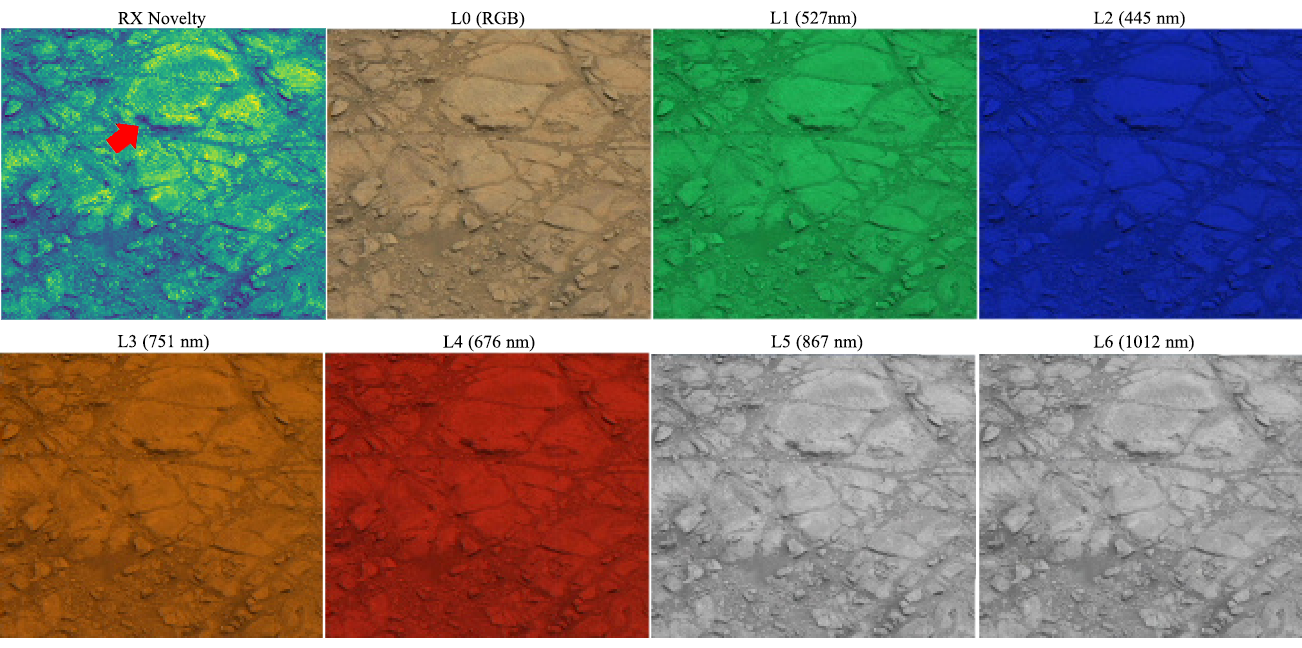}
\caption{\bf{An example of difficult to notice novel features from mcam12276. In this example, there is a subtle ring of brighter dust on the edge of one of the rocks (shown by the red arrow). This is not noticeable in L0 (RGB) and faint in L5 (867 nm) and L6 (1012nm).}}
\label{mcam12276}
\end{figure*}

To assist tactical operations planning, the novelty detection system is most useful when it is able to highlight features of interest that are not otherwise easily identifiable. 
Figure \ref{mcam12276} shows an example of such a sequence.
In this sequence, there is a ring of novel material around one of the rocks near the top of the thumbnail. 
This ring is almost unidentifiable in the RGB image and faint in filters L5 and L6. 
This ring is suspected by science team members to be due to a thin layer of brighter regolith deposited around the edge of the rock and not part of the rock composition.
By taking every filter into account, RX is able to identify the spectral anomaly around this rock and provide guidance for follow up analysis.

\begin{figure*}
\centering
\includegraphics[width=\linewidth]{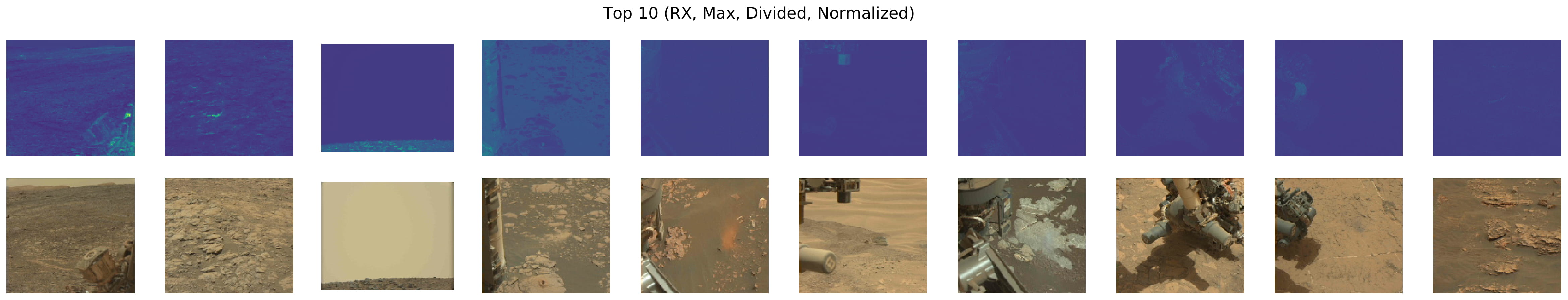}
\includegraphics[width=\linewidth]{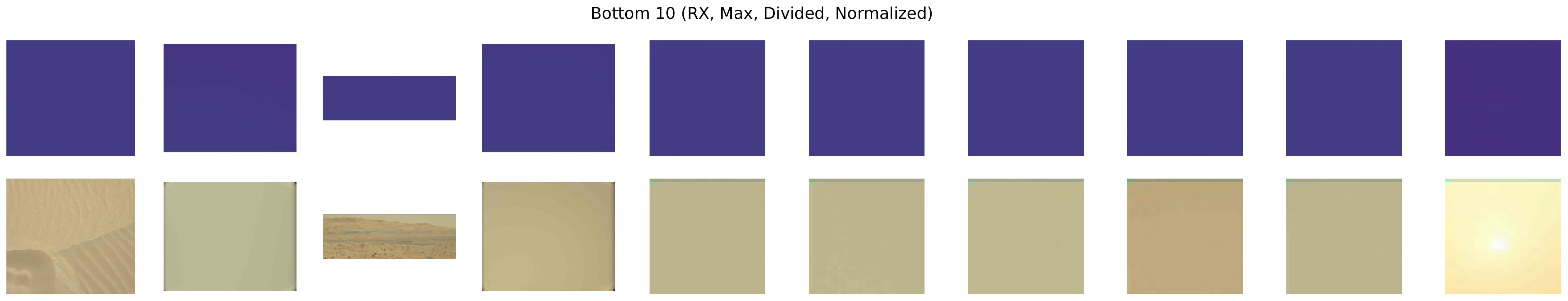}
\caption{\bf{The most and least novel sequences in the Mastcam left eye data set. Sequences are sorted by maximum novelty value obtained from a brightness corrected background distribution. The pixel novelty scores  are normalized across the entire data set.}}
\label{sorted}
\end{figure*}

In order to assess which sequences to analyze first, statistics about the pixel novelty scores can be calculated to rank multispectral sequences. 
Sorting by the mean pixel novelty score in each sequence orders the sequences based on mean novelty across the image.
%While the most novel sequences will rise to the top of the stack, ordering by mean pixel novelty score provides unexpected results. 
However, sorting by the mean score of all pixels in the sequence does not effectively prioritize images with obvious or localized novel features because the mean score can suppress high scores is small, localized features that are the only novel feature in the image. 
%Sequences with a high average novelty score typically consist of images without any obvious novelties as the whole image has a high average score. 
In contrast, images with high mean scores that have uniformly high scores across the entire image will not show contrast in the heat map visualization and will be difficult to interpret using this visualization. 
%This type of sorting is not fitting for tactical operations analysis as the heat map provides little useful information. 
To prioritize high novelty scores corresponding to localized features, we used the maximum pixel novelty score instead to sort novel sequences as it is better at prioritizing images with high novelty features. 
%Maximum pixel novelty score highlights sequences with clear outliers making them clearer candidates for follow up analysis. 
Examples of the top and bottom ten images sorted by maximum RX score are shown in Figure \ref{sorted}. 
The top-scoring sequences show images of high variability when compared with the low-scoring images. 
The highest scoring pixel in these sequences are typically rover parts which, while being spectrally novel, are not necessarily useful for operations.
Additionally, sequences containing shadows that move over time tend to have higher scores, as the shadow fans across the different channels and creates a rainbow.
The low-scoring sequences mostly consist of sky and sand images which are relatively low interest from a tactical perspective. 
When comparing multiple images, we normalized the pixel novelty scores across the data set in order to display relative novelty. 
The scores are visualized without normalization when an analyst selects an individual sequence to analyze. 

%%%%%%%%%%%%%%%%%%%%%%%%%%%%%%%%%%%%%%
\section{Conclusions}
%%%%%%%%%%%%%%%%%%%%%%%%%%%%%%%%%%%%%%
While initial results from reviewing the novelty detection output seem promising, it is difficult to evaluate the effectiveness of the system at assisting in operations without a longer term study in which many sequences would be analyzed for previously undiscovered features.
This is necessary as generating a test set of undiscovered novelties is not feasible.
%tell the percentage of new novelties present. 
%As this study is looking for hard to detect novelties, generating a labeled test set for the undiscovered is not feasible.
The purpose of this study is to assist tactical planning so the key value is in the system's ability to highlight unique sequences and novel features within those sequences. 
The brightly colored heat map of novel regions for each sequence appears to provide a fast way of identifying spots for follow up analysis. 
The next steps for this work are to gather more feedback from the tactical planning team to determine the productivity increase while using novelty detection products.
To quantitatively evaluate this, operations staff members can be given the task of analyzing Mastcam sequence with and without the novelty-based products.
Analysis time can be recorded to identify if staff members are able to identify novel features quicker when the novelty detection system is used for triaging.
Instances where the system highlights something that the staff member does not find novel can also help to understand the limitations of the method and improve future method development.
This feedback will help to validate the usability of the novelty detection system and fine tune the features of the system, such as dividing by the brightness, global normalization, and limiting the training set to recent data. 
Additionally, the novelty detection products may improve masking rover parts and finding ways to reduce the rainbowing effect from shadows.
Another approach may be to not use a background distribution based on the entire Mastcam multispectral history and instead calculating new models for each image to find the most novel feature on a local scale.

Future work is also needed to improve ways of ordering the sequences based on their novelty scores.
While sorting sequences by the maximum pixel RX score provides more useful outputs than sorting by the mean pixel RX score, there may be better ways of prioritizing sequences (e.g., sorting by the variance of scores in the image). %heat map may help to identify sequences with spectral novelties.
To validate that the ranking's priorities align with the triage an operations staff member may perform, an experiment can also be created to have both humans and the novelty detection system generate rankings.
The rankings can be compared using Spearman's rank correlation coefficient to compare their overlap.
Such an experiment could also help to find better ways of sorting the sequences.
Finding better ways to prioritize novel sequences is also beneficial for on-board autonomy applications \cite{wagstaffnovelty}.

MSLWEB, an Arizona State University-based Mastcam tracking application, is the perfect platform for integrating novelty detection products into tactical planning workflows. 
Currently, MSLWEB supports the automatic generation of simple quick look products such as decorrelation stretches and filter ratios. 
Augmenting this system to support novelty detection would involve adding an inference endpoint to the pipeline and automatically passing new sequences through it. 
As this system is currently in use by operations staff, a strong justification is necessary to perform this integration. 
At the time, the novelty detection system has shown promising examples of novelty highlighting and sorting. 
Further analysis is needed to demonstrate how these examples would fit into the tactical planning pipeline and the time saving benefits it might provide.

%%%%%%%%%%%%%%%%%%%%%%%%%%%%%%%%%%%%%%%%%%%%%%%%%%%%%%%%%%%%%%%%%%%%%%%%%%%%%%%%%%%%%%%%%%%%%%%%%%%%%%
\acknowledgments
The authors thank the NASA Jet Propulsion Laboratory (JPL) Strategic University Research Partnership (SURP), the JPL Visiting Student Researcher Program (JVSRP), and the NASA Space Technology Graduate Research Opportunity (NSTGRO) for supporting this project.  
The research was carried out in part at the Jet Propulsion Laboratory, California Institute of Technology, under a contract with the National Aeronautics and Space Administration (80NM0018D0004).

%%%%%%%%%%%%%%%%%%%%%%%%%%%%%%%%%%%%%%%%%%%%%%%%%%%%%%%%%%%%%%%%%%%%%%%%%%%%%%%%%%%%%%%%%%%%%%%%%%%%%%
\bibliographystyle{IEEEtran}
\bibliography{ieee}

%%%%%%%%%%%%%%%%%%%%%%%%%%%%%%%%%%%%%%%%%%%%%%%%%%%%%%%%%%%%%%%%%%%%%%%%%%%%%%%%%%%%%%%%%%%%%%%%%%%%%%
\thebiography 
%% This biostyle allows you to insert your photo size 1in X 1.25in
\begin{biographywithpic}
{Paul Horton}{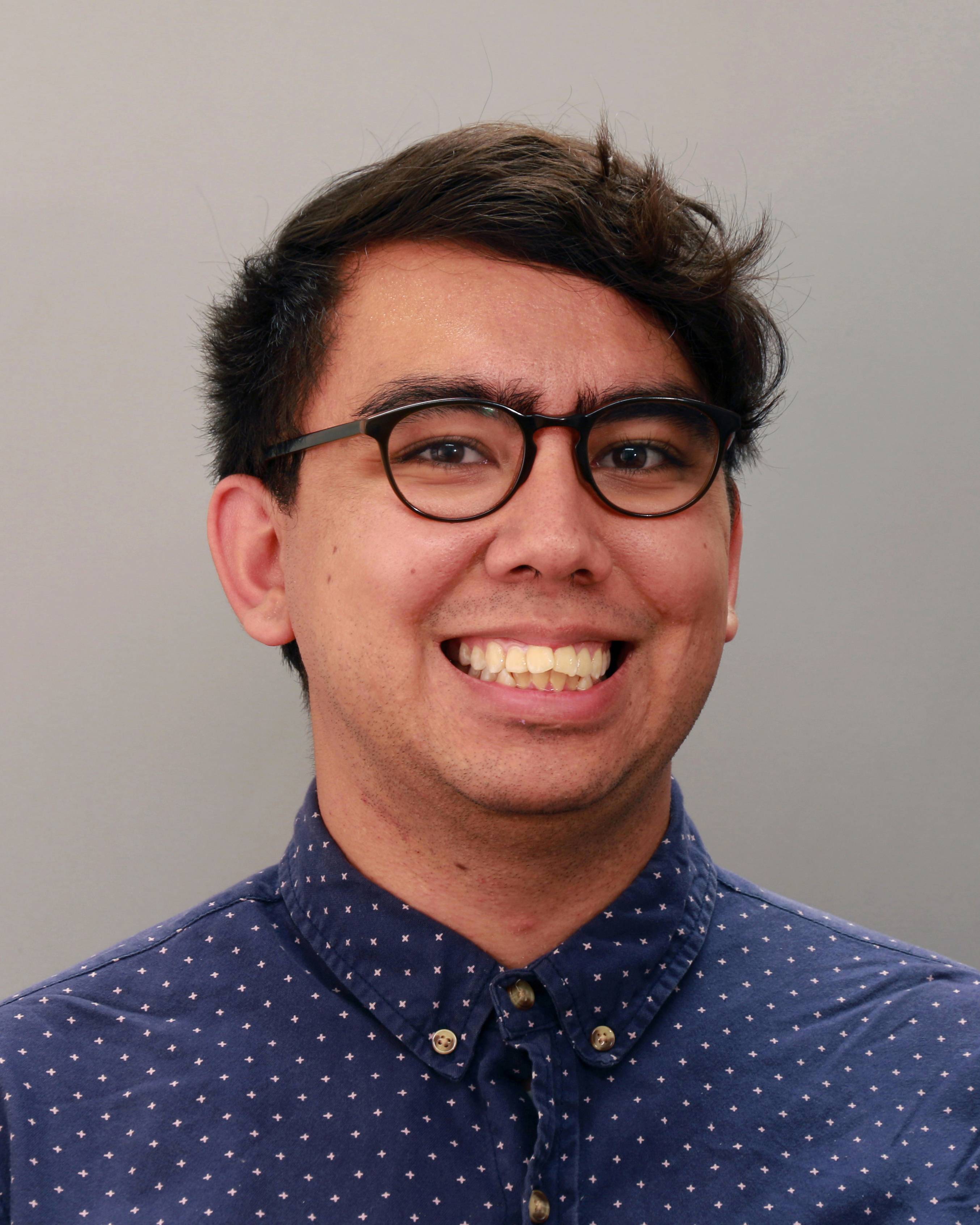} 
received B.S. degrees in Applied Physics and Software Engineering in 2018, his M.S. in Software Engineering in 2019, and is pursuing his Ph.D. in Exploration Systems Design (Systems Engineering) at Arizona State University.
He is a recipient of the NASA Space Technology Graduate Research Opportunity (NSTGRO) for his work on integrating data science systems into planetary science and astronomy and plans to graduate in 2023.
He works closely with the Machine Learning and Instrument Autonomy Group at the Jet Propulsion Laboratory, California Institute of Technology, on multi-spectral anomaly detection with Curiosity and on-board autonomy for balloon operations.
\end{biographywithpic}

\begin{biographywithpic}
{Hannah R. Kerner}{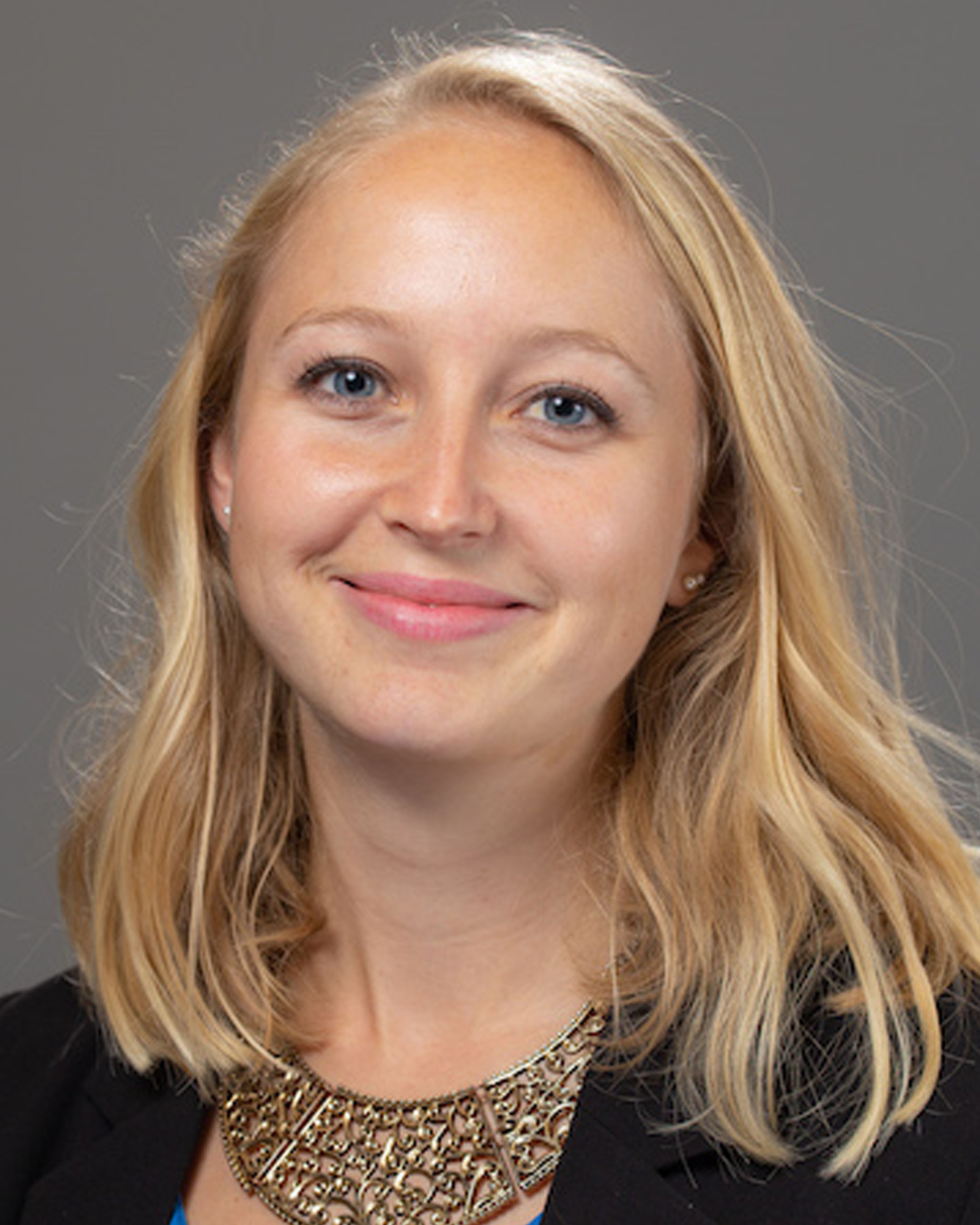} 
received her B.S. degree from the University of North Carolina at Chapel Hill and her Ph.D. degree from Arizona State University. She is an Assistant Research Professor at the University of Maryland College Park where her research focuses on developing new machine learning methods for remote sensing applications. She leads Machine Learning and U.S. Domestic projects in the NASA Harvest program. 
\end{biographywithpic}

\begin{biographywithpic} 
{Samantha Jacob}{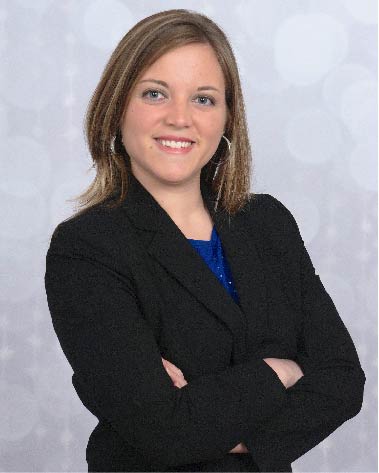}
is a PhD candidate working with Dr. Jim Bell at Arizona State University studying the
mineralogy of Mars using multispectral data. She has her B.S. and M.S. in Geology and Geophysics from
the University of Hawaii at Manoa. It was during her masters in 2012 that she started working with data
from Mars rovers, working first with data from the MSL Curiosity rover. While still in charge of
calibrating data from the Curiosity Mastcam instrument, she has also joined the Mastcam-Z team at ASU
to prepare the calibration procedures for the Mars 2020 Perseverance rover.
\end{biographywithpic}

\begin{biographywithpic}
{Ernest Cisneros}{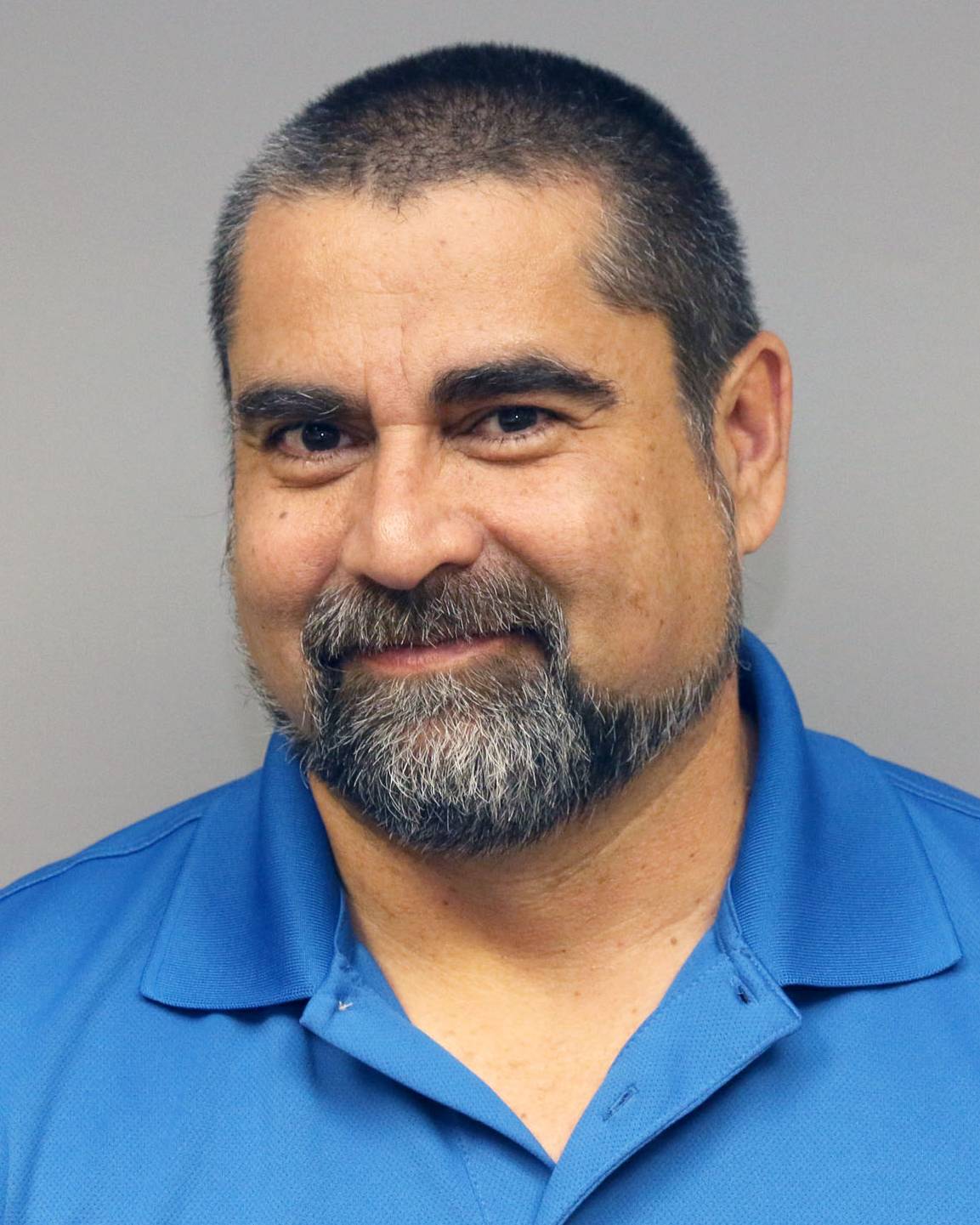}
received his B.S degree in Geology from University of Texas at El Paso. Ernest leads the Science Data Center, within  the School of Earth and Space Exploration, which supports instrument and spacecraft operations, and science data analysis programs. The SDC currently supports Mastcam (MSL), Mastcam-Z (M2020), Multispectral Imager (Psyche) and TTCAM (Lucy) instruments, and the LunaH-Map cubesat spacecraft. Prior to joining ASU in 2007, he worked at Northwestern University, Duke University and the U.S. Geological Survey, providing programming, network and system administration support to academic and research programs.
\end{biographywithpic} 

\begin{biographywithpic}
{Kiri L. Wagstaff}{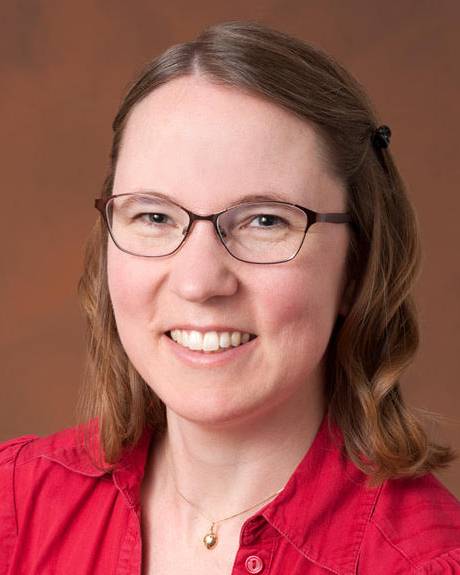}
received a B.S. degree (1997) in Computer Science from the University of Utah, M.S. (2000) and Ph.D. (2002) degrees in Computer Science from Cornell University, an M.S. degree (2008) in Geological Sciences from the University of Southern California, and an MLIS degree (2017) in Library and Information Science from San Jose State University.
She is a Principal Research Technologist in artificial intelligence and machine learning at the Jet Propulsion Laboratory and an Associate Research Professor at Oregon State University. Her research focuses on developing new machine learning and data analysis methods for spacecraft.
\end{biographywithpic}

\begin{biographywithpic}
{Jim Bell}{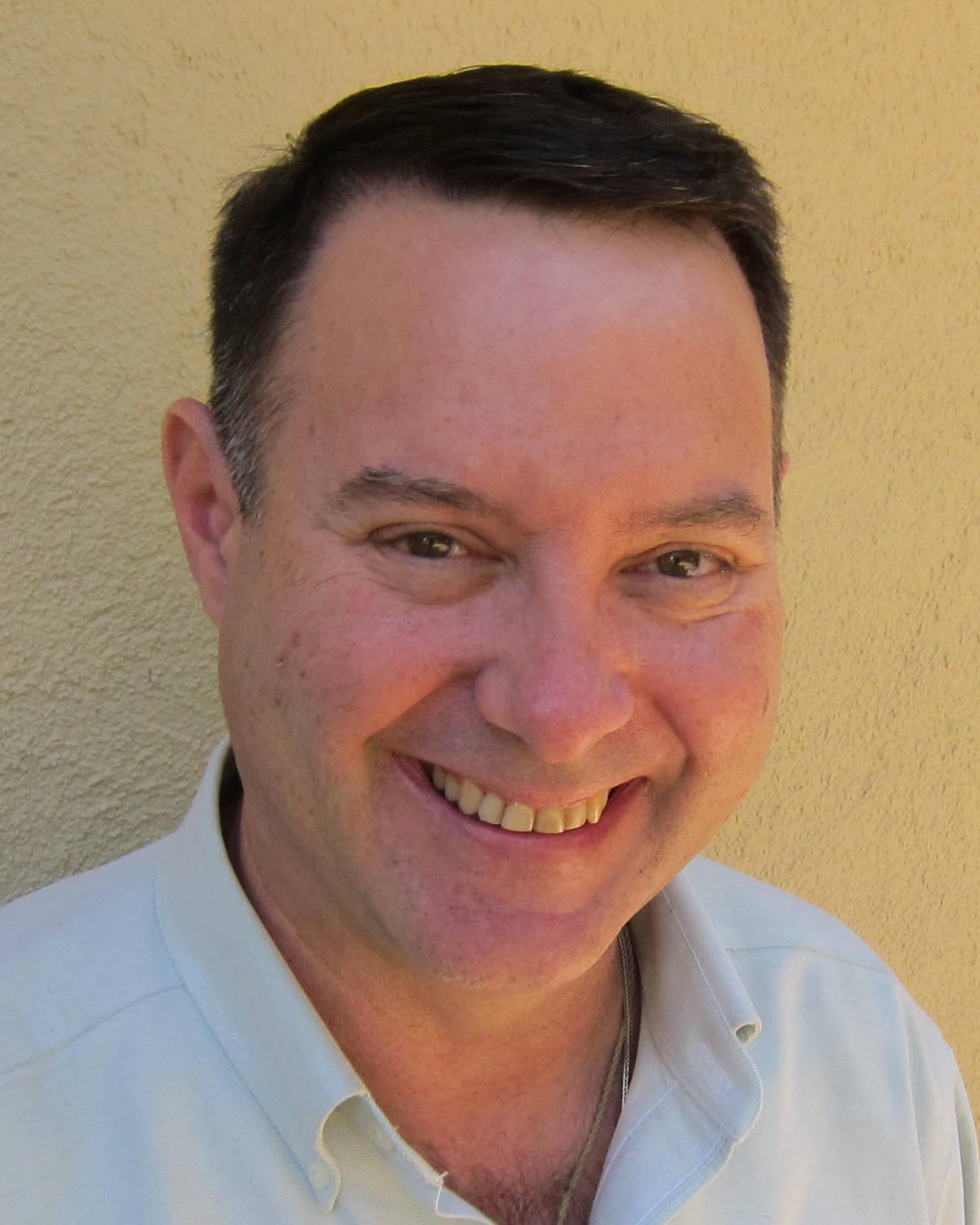}
is a professor in the School of Earth \& Space Exploration at Arizona State University, where he teaches courses in astronomy, geology, planetary science, and commercial space. He is an active astronomer and planetary scientist who has been involved in solar system exploration using the Hubble Space Telescope, Mars rovers, and orbiters sent to Mars, the Moon, and several asteroids. His research focuses on the use of remote sensing imaging and spectroscopy to assess the geology, composition, and mineralogy of the surfaces of planets, moons, asteroids, and comets. He is also the author of many popular science books related to space exploration.
\end{biographywithpic}

\end{document}